\documentclass[10pt,twocolumn,letterpaper]{article}

\usepackage{iccv}
\usepackage{times}
\usepackage{epsfig}
\usepackage{graphicx}
\usepackage{amsmath}
\usepackage{amssymb}
\usepackage{caption}
\usepackage{bm}
\usepackage{booktabs}
\usepackage[font=small]{caption}
\usepackage{authblk}

\usepackage[breaklinks=true,bookmarks=false]{hyperref}

\usepackage{xcolor}

\iccvfinalcopy

\begin{document}

\title{Instruct-NeRF2NeRF: Editing 3D Scenes with Instructions}

\author{Ayaan Haque}
\author{Matthew Tancik}
\author{Alexei A. Efros}
\author{Aleksander Holynski}
\author{Angjoo Kanazawa}
\affil{UC Berkeley}

\twocolumn[{
\renewcommand\twocolumn[1][]{#1}%
\maketitle
\begin{center}
    \captionsetup{type=figure}
    \includegraphics[width=\linewidth]{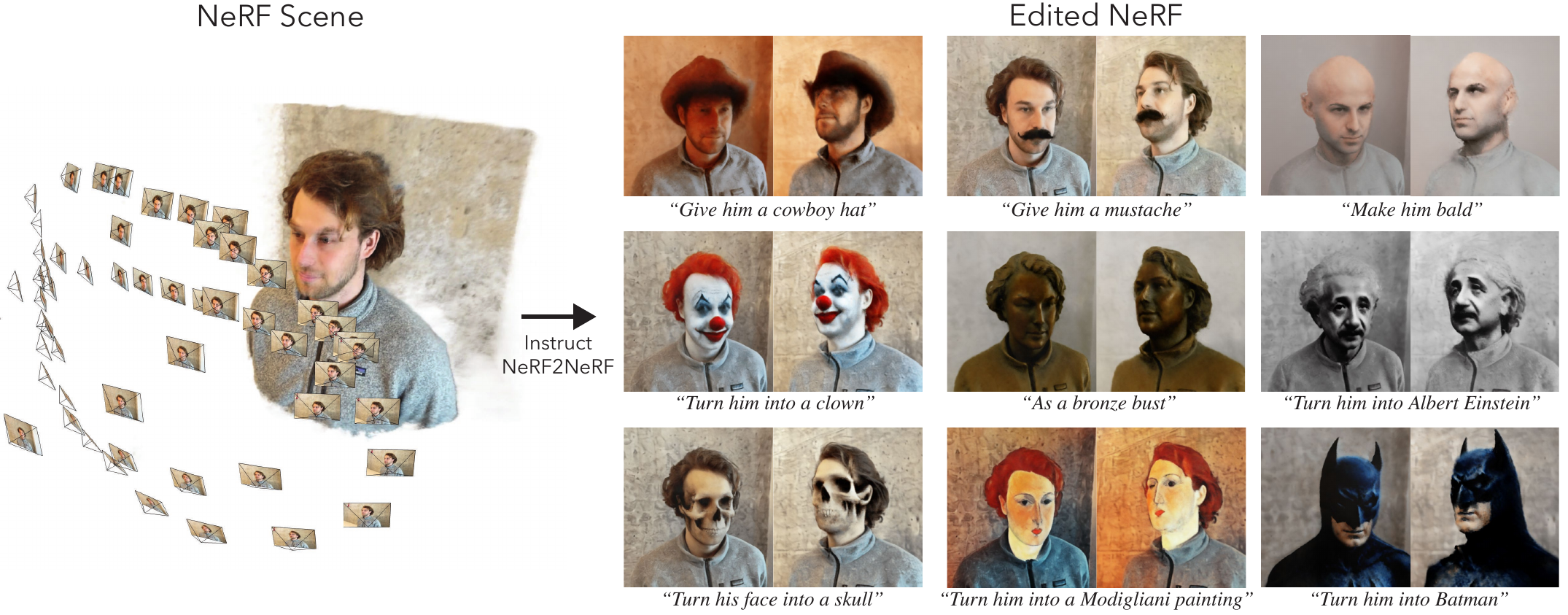}
    \captionof{figure}{\small \textbf{Editing 3D scenes with Instructions.} We propose Instruct-NeRF2NeRF, a method for consistent 3D editing of a NeRF scene using text-based instructions. Our method can accomplish a diverse collection of local and global scene edits.}
    \label{fig:teaser}
\end{center}

}]

\begin{abstract}
   We propose a method for editing NeRF scenes with text-instructions. Given a NeRF of a scene and the collection of images used to reconstruct it, our method uses an image-conditioned diffusion model (InstructPix2Pix) to iteratively edit the input images while optimizing the underlying scene, resulting in an optimized 3D scene that respects the edit instruction. We demonstrate that our proposed method is able to edit large-scale, real-world scenes, and is able to accomplish more realistic, targeted edits than prior work. Result videos can be found on the project website: \url{https://instruct-nerf2nerf.github.io}.
\end{abstract}

\section{Introduction}

\begin{figure*}
    \centering
    \includegraphics[width=.9\linewidth]{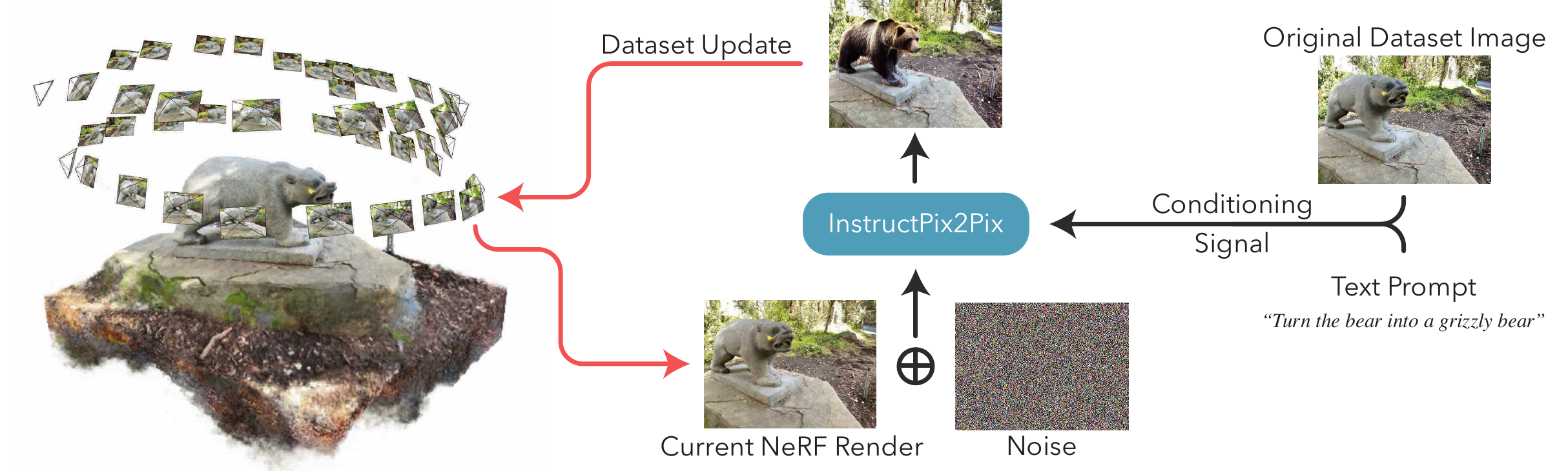}
    \caption{\textbf{Overview}: Our method gradually updates a reconstructed NeRF scene by iteratively updating the dataset images while training the NeRF: (1) an image is rendered from the scene at a training viewpoint, (2) it is edited by InstructPix2Pix given a global text instruction, (3) the training dataset image is replaced with the edited image, and (4) the NeRF continues training as usual.}
    \label{fig:pipeline}
\end{figure*}

With the emergence of efficient neural 3D reconstruction techniques, capturing a realistic digital representation of a real-world 3D scene has never been easier. The process is simple: capture a collection of images of a scene from varying viewpoints, reconstruct their camera parameters, and use the posed images to optimize a Neural Radiance Field~\cite{mildenhall2020nerf}. Due to its ease of use, we expect captured 3D content to gradually replace the traditional processes of manually-generated assets. Unfortunately, while the pipelines for turning a real scene into a 3D representation are relatively mature and accessible, many of the other necessary tools for the creation of 3D assets (e.g., those needed for \emph{editing} 3D scenes) remain underdeveloped. 

Traditional processes for editing 3D models involve specialized tools and years of training in order to manually sculpt, extrude, and re-texture a given object. This process is made even more involved with the advent of neural representations, which often do not have explicit surfaces. This further motivates the need for 3D editing approaches designed for the modern era of 3D representations, particularly approaches that are similarly as accessible as the capture techniques themselves.

\looseness=-1 To this end, we propose Instruct-NeRF2NeRF, a method for editing 3D NeRF scenes that requires as input only a text instruction. Our approach operates on a pre-captured 3D scene and ensures that the resulting edits are reflected in a 3D-consistent manner. For example, given a 3D scene capture of a person shown in Figure~\ref{fig:teaser} (left), we can enable a wide variety of edits using flexible and expressive textual instruction such as \textit{``Give him a cowboy hat''} or \textit{``Turn him into Albert Einstein''}.
Our approach makes 3D scene editing accessible and intuitive for everyday users.

\looseness=-1 Though there exist 3D generative models, the datasources required for training these models at scale are still limited. Therefore, we instead choose to extract shape and appearance priors from a 2D diffusion model. Specifically, we employ a recent image-conditioned diffusion model, InstructPix2Pix~\cite{brooks2022instructpix2pix}, which enables instruction-based 2D image editing. Unfortunately, applying this model on individual images rendered from a reconstructed NeRF produces inconsistent edits across viewpoints. As a solution to this, we devise a simple approach similar to recent 3D generation solutions like DreamFusion~\cite{poole2022dreamfusion}. Our underlying method, which we refer to as Iterative Dataset Update (Iterative DU), alternates between editing the ``dataset'' of NeRF input images, and updating the underlying 3D representation to incorporate the edited images.

We evaluate our approach on a variety of captured NeRF scenes, validating our design choices by comparing with ablated variants of our method, as well as na\"ive implementations of the score distillation sampling (SDS) loss proposed in DreamFusion~\cite{poole2022dreamfusion}.
We also qualitatively compare our approach to a concurrent text-based stylization approach~\cite{wang2022nerf}.
We demonstrate that our method can accomplish a wide variety of edits on people, objects, and large-scale scenes.

\section{Related Work}

\paragraph{Physical Editing of NeRFs}
NeRFs~\cite{mildenhall2020nerf} are a popular approach for generating photorealistic novel views of a scene captured by calibrated photographs and have been extended in many follow-up works~\cite{tewari2022advances}. 
However, editing NeRFs remains a challenge due to their underlying representation. One approach is to impose physics-based inductive biases in its optimization process to enable changes in materials or scene lighting
~\cite{verbin2022refnerf,boss2021neuralpil,nerv2021,munkberg2022extracting,mildenhall2022rawnerf}.
Alternatively, one can specify bounding boxes~\cite{ost2021neural,orf}, to allow easy compositing of different objects~\cite{zhang2021editable} as well as spatial manipulations and geometry deformations~\cite{yuan2022nerf}. A recent work, ClimateNeRF~\cite{li2022climatenerf}, extracts rough geometry from a NeRF and uses physical simulation to apply weather changes such as snowing and flooding. 
Most physically-based edits revolve around changing physical properties of the reconstructed scene, or performing physical simulation. In this work, we instead focus on enabling arbitrary creative edits. 

\paragraph{Artistic Stylization of NeRFs}
Following the literature from image stylization~\cite{hertzmann1998painterly,gatys2016image}, recent works have explored artistic 3D stylization of NeRFs~\cite{chiang2021stylizing,huang_2021_3d_scene_stylization,huang2022stylizednerf,nguyen2022snerf,zhang2022arf,wu2022palettenerf}. While these approaches can obtain 3D-consistent stylizations of a scene, they primarily focus on global scene appearance changes and usually require a reference image. Other works have explored the use of latent representations from visual language models such as CLIP~\cite{radford2021learning}. EditNeRF~\cite{liu2021editing} explores editing NeRFs by manipulating latent codes learned from object categories in a synthetic dataset.  To increase usability (and as explored in other 3D domains such as meshes~\cite{text2mesh,hong2022avatarclip}), ClipNeRF~\cite{wang2021clip} and NeRF-Art~\cite{wang2022nerf} extend this line of work by encouraging similarity between CLIP embeddings of the scene and a short text prompt.
A limitation of these CLIP-based approaches is their inability to incorporate localized edits.
Methods such as Distilled Feature Fields~\cite{kobayashi2022distilledfeaturefields} and Neural Feature Fusion Fields~\cite{tschernezki22neural} distill 2D features from pre-trained models such as LSeg~\cite{li2022language} and DINO~\cite{caron2021emerging} into the radiance fields, which enable specification of regions. These approaches allow for localized CLIP-guided edits, 3D spatial transformations~\cite{wang2021clip}, or localized scene removal~\cite{tschernezki22neural} specified either by language or a reference image. In this work, we offer a complementary approach to editing 3D scenes based on intuitive, purely language-based editing instructions. While masking enables specific local changes, instructional edits provide intuitive high-level instructions that can make more flexible and holistic changes to the appearance or geometry of a single object or the entire scene. We enable mask-free instructional edits by taking advantage of recent instruction-based 2D image-conditioned diffusion model~\cite{brooks2022instructpix2pix}, resulting in a purely-language-based interface that enables a wider range of intuitive and content-aware 3D editing. 

\paragraph{Generating 3D Content}
Recent progress in pre-trained large-scale models has enabled rapid progress in the domain of generating 3D content from scratch, either by optimizing radiance fields through vision-language models like CLIP~\cite{jain2021dreamfields,lee2022understanding} or via text-conditioned diffusion models~\cite{ramesh2022hierarchical,saharia2022photorealistic,rombach2022high} as presented in DreamFusion~\cite{poole2022dreamfusion} and its follow-ups~\cite{wang2022score,lin2022magic3d,metzer2022latent}. 
While these approaches can generate 3D models from arbitrary text prompts, they lack (1) fine-grained control over the synthesized outputs, (2) the ability to generalize to scenes (\ie, anything beyond a single object isolated in space), and (3) any grounding in reality, producing entirely synthesized creations. Concurrent works such as RealFusion~\cite{melas2023realfusion} and SparseFusion~\cite{zhou2022sparsefusion} explore grounding by providing one or few input images, where the unseen parts are hallucinated. In all of these approaches, a central challenge is congealing the inconsistent outputs of a 2D diffusion model into a consistent 3D scene. 
In this work, instead of creating new content, we focus on editing  \emph{real} captured NeRFs of \emph{fully observed scenes} using 2D diffusion priors. One advantage of editing an existing NeRF scene (as opposed to generating 3D content from scratch) is that the captured images are by definition 3D consistent, suggesting that generated imagery should naturally be more consistent. This also helps avoid certain design decisions that result in the cartoon-ish appearance commonly seen in unconditional 3D content generation methods~\cite{poole2022dreamfusion,wang2022score,lin2022magic3d}.

\paragraph{Instruction as an Editing Interface}
With the rise of large-language models (LLMs) like GPT~\cite{brown2020language} and Chat-GPT~\cite{chatgpt}, natural language is emerging as the next ``programming language'' for specifying complex tasks.
LLMs allow for the abstraction of a series of low-level specifications into an intuitive and user-friendly interface through the use of language, specifically \textit{instructions}~\cite{ouyang2022training}. InstructPix2Pix~\cite{brooks2022instructpix2pix} demonstrates the effectiveness of instructions in 2D image tasks, as do other works in other domains such as robotic navigation~\cite{huang23vlmaps}.
We propose the first work that demonstrates instructional guidance in the realm of 3D editing. 
This is particularly significant given the difficulty of the task, which has typically required specialized tools and years of experience. 
By using natural language instructions, even novice users can achieve high-quality results without additional tools or specialized knowledge.

\begin{figure}
    \centering
    \includegraphics[width=\linewidth]{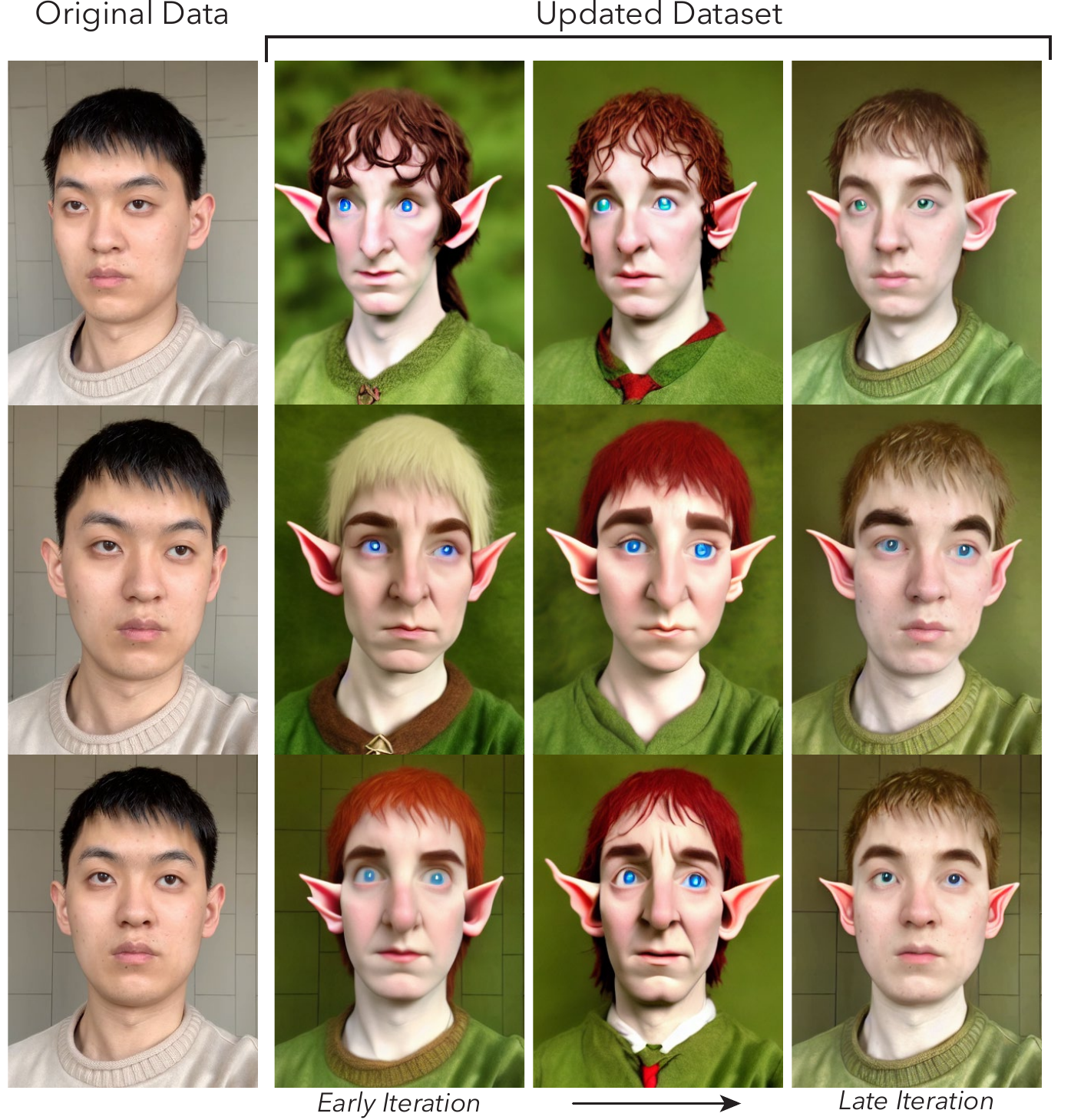}
    \caption{\textbf{Dataset Evolution}: At the start of training, the edited images perform the requested edit, but are often inconsistent. After iteratively training the NeRF and updating the training dataset, the images gradually become more 3D consistent.}
    \label{fig:dataset_evolution}
\end{figure}

\begin{figure*}
    \centering
    \includegraphics[width=\linewidth]{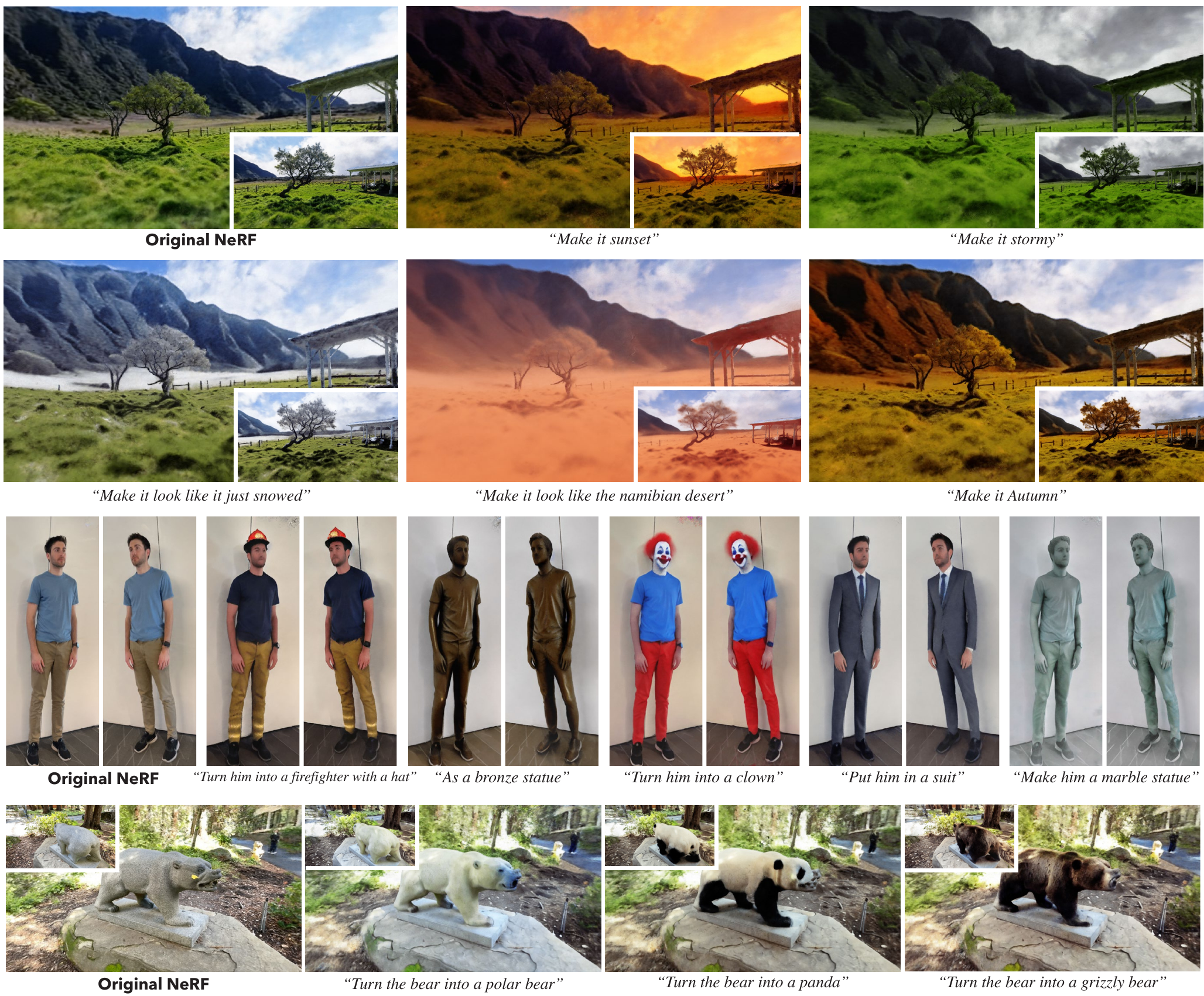}
    \caption{\textbf{Qualitative Results:} Our method is able to perform a variety of diverse contextual edits on real scenes, including environmental changes, like adjusting the time of day, and even more localized changes that modify only a specific object in the scene.}
    \label{fig:main_results}
\end{figure*}

\section{Method}
Our method takes as input a reconstructed NeRF scene along with its corresponding source data: a set of captured images, their corresponding camera poses, and camera calibration (typically from a structure-from-motion system, such as COLMAP~\cite{schonberger2016structure}). Additionally, our method takes as input a natural-language editing instruction, e.g., \emph{``turn him into Albert Einstein"}. As output, our method produces an edited version of the NeRF subject to the provided edit instruction, as well as edited versions of the input images.

Our method accomplishes this task by iteratively updating the image content at the captured viewpoints with the help of a diffusion model, and subsequently consolidating these edits in 3D through standard NeRF training. Our work builds off recent advances in diffusion models for image editing, specifically InstructPix2Pix~\cite{brooks2022instructpix2pix}, which proposes an image-and-text conditioned diffusion model trained to edit natural images using human-provided instructions.

\subsection{Background}
\paragraph{Neural radiance fields}
\label{sec:nerf_prelim}
Neural radiance fields (NeRFs)~\cite{mildenhall2020nerf} are a compact and convenient representation for reconstructing and rendering a volumetric 3D scene. A NeRF is parameterized by 3D positions $(x,y,z)$ and viewing directions $(\theta,\phi)$ for samples in a field. Each sample is processed to produce a color and density $(\bm c, \sigma)$, which can be composited along a ray to produce a 2D pixel color. A NeRF is optimized using a collection of captured images and their corresponding camera parameters, which include both calibration and extrinsic pose/orientation. These camera parameters can be used to extract a per-pixel world-space ray parameterization that describes the 3D center $\bm o$ and direction $\bm d$ of the camera ray ${\bm r}(t) = {\bm o} + t{\bm d}$ that corresponds to each pixel in each image. These rays with their associated pixel colors are used to optimize the NeRF. The typical process of training a NeRF \cite{mildenhall2020nerf} involves selecting a subset of rays $\bm r$, rendering the NeRF's current estimate of the color along this ray $\hat{C}(r)$, and computing a loss relative to captured pixel color $\mathcal{L}(C(r), \hat{C}(r))$. In practice, in the interest of reliable optimization, rays are selected at random from a variety of viewpoints, to ensure the 3D positions of reconstructed scene objects are sufficiently well-constrained. To render a novel viewpoint, a collection of rays are sampled corresponding to all the pixels in that novel image, and the resulting color values $\hat{C}(r)$ are arranged into a 2D frame.

\paragraph{InstructPix2Pix}
\label{sec:ip2p_prelim}

\looseness=-1 Denoising diffusion models~\cite{sohl2015deep, ho2020denoising} are generative models that learn to gradually transform a noisy sample towards a modeled data distribution.
InstructPix2Pix~\cite{brooks2022instructpix2pix} is a diffusion-based method specialized for image editing. Conditioned on an RGB image $c_I$ and a text-based editing instruction $c_T$, and taking as input a noised image (or pure noise) $z_t$, the model aims to produce an estimate of the edited image $z_0$ (an edited version of $c_I$ subject to the instruction $c_T$). Formally, the diffusion model predicts the amount of noise present in the input image $z_t$, using the denoising U-Net $\epsilon_\theta$ as:
\begin{equation}
\hat{\epsilon} = \epsilon_\theta(z_t; t, c_I, c_T)
\end{equation}
This noise prediction $\hat{\epsilon}$ can be used to derive $\hat{z}_0$, the estimate of the edited image. This denoising process can be queried with a noisy image $z_t$ at any timestep $t\in[0,T]$, i.e., containing any amount of noise, up to a pure noise image $z_T$. Larger amounts of noise, i.e., larger values of $t$, will produce estimates of $\hat{z}_0$ with more variance, whereas smaller $t$ values will produce lower variance estimates with more adherence to the visible image signal in $z_t$. 

In practice, InstructPix2Pix is based on a latent diffusion model~\cite{rombach2022high}, i.e., the diffusion process operates entirely on an encoded latent domain. This means that the above-defined variables $c_I, z_0$ are all latent images created by encoding an RGB image, i.e., $\mathcal{E}(I)$. Similarly, to produce an RGB image from the diffusion model, one must also decode the predicted $\hat{z}_0$ latents via the decoder $\hat{I} = \mathcal{D}(\hat{z}_0)$.

\begin{figure}
    \centering
    \includegraphics[width=\linewidth]{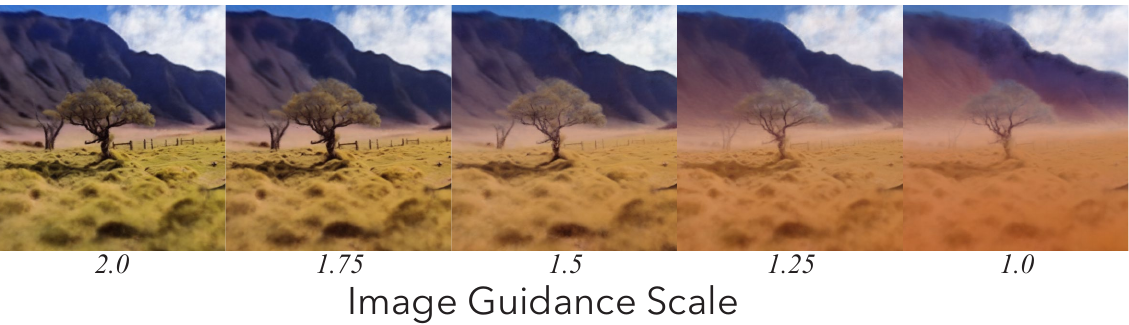}
    \caption{\textbf{Guidance Scale:} By varying the image guidance we can control how much the edit looks like the original scene. Note that these are renderings from the edited 3D scenes.}
    \label{fig:cfg_Scale}
\end{figure}
\subsection{Instruct-NeRF2NeRF}

\label{sec:in2n}

Given a reconstructed NeRF scene (including the corresponding dataset of calibrated images), as well as a text instruction, we fine-tune the reconstructed model towards an edit instruction to produce an edited version of that NeRF. An overview is provided in Fig.~\ref{fig:pipeline}.

Our method works through an alternating update scheme, in which the training dataset images are iteratively updated using a diffusion model and are subsequently consolidated into the globally consistent 3D representation by training the NeRF on these updated images. This iterative process allows for gradual percolation of the diffusion priors into the 3D scene. Although this process can enable significant edits to the scene, our use of an image-conditioned diffusion model (InstructPix2Pix) helps in maintaining the structure and identity of the original scene.

In this section, we first describe our use of InstructPix2Pix in the process of editing a single dataset image, then describe our iterative procedure for gradually updating dataset images and refining the reconstructed NeRF. 

\paragraph{Editing a rendered image}

We use InstructPix2Pix~\cite{brooks2022instructpix2pix} to edit each dataset image. It takes three inputs: (1) an input conditioning image $c_I$, a text instruction $c_T$, and a noisy input $z_t$. To update a dataset image at viewpoint $v$, we use the unedited image $I_0^v$ for $c_I$, which will typically be the originally captured image at this viewpoint, or if the viewpoint was not captured physically, a render from the NeRF before any edits were made. For $z_t$, as in SDEdit~\cite{meng2021sdedit}, we input a noised version of the current render at optimization step $i$, i.e., a linear combination of $\mathcal{N}(0,1)$ and $z_0 = \mathcal{E}(I_i^v)$. For simplicity, we denote the process of replacing an image $I_i^v$ as $I_{i+1}^v\leftarrow U_\theta(I_{i}^v, t; I_0^v, c_T)$, where a noise level $t$ is chosen at random from a constant range $[t_\text{min}, t_\text{max}]$. We define $U_\theta$ as the DDIM sampling process, with a fixed number of intermediate steps $s$ taken between initial timestep $t$ and $0$.

This process mixes two sources of information: the diffusion model aims to edit the original image $I_0^v$ according to the instruction $c_T$, while the noised image passed to the diffusion U-Net $z_t$ is only partially noised (with some $t<T$), such that the rendering of the current global 3D model influences the diffusion model's final estimate for $z_0$ (the image which will replace the dataset image at viewpoint $v$). A key thing to note is that while our method continually repeats the process of rendering from the NeRF, editing the image, and updating the NeRF, the diffusion model is conditioned on the \emph{un-edited} images, and thus remains grounded, preventing the characteristic drift commonly associated with recurrent synthesis. 

\paragraph{Iterative Dataset Update}

The core component of our method is an alternating process through which images are rendered from the NeRF, updated by the diffusion model, and subsequently used to supervise the NeRF reconstruction. We refer to this process as the Iterative Dataset Update (\emph{Iterative DU}).

When optimization begins, our image dataset consists of the originally captured images from a range of viewpoints denoted as $v$, which we represent as $I_0^v$. These images are cached separately and used as conditioning for the diffusion model at all stages. At each iteration, we perform a number of image updates $d$, followed by a number of NeRF updates $n$. Image updates are performed sequentially in a random ordering of $v$ determined at the start of training.
NeRF updates always sample a set of random rays from the entire training dataset, such that the supervision signal contains a mixture of \textit{old} information and pixels from recently updated dataset images. 

Our editing process results in sudden replacement of dataset images with their edited counterpart. At early iterations, these images may perform inconsistent edits (as InstructPix2Pix does not typically perform consistent edits across different viewpoints). Over time, as images are used to update the NeRF and progressively re-rendered and updated, they begin to converge on a globally consistent depiction of the edited scene. Examples of this evolution process can be seen in Figure~\ref{fig:dataset_evolution}. 

This process is similar to the approach proposed in SNeRF \cite{nguyen2022snerf}, where the images are updated through style transfer in every alternate iteration. Unlike SNeRF, our iterative DU retains edited images across NeRF updates, effectively performing semi-permanent updates to the training dataset. This process can also be interpreted as a variant of the score distillation sampling (SDS) loss from DreamFusion~\cite{poole2022dreamfusion}, where instead of updating a discrete set of images at each step, each gradient update contains a random mixture of rays distributed across many viewpoints, and the computed gradients along these rays may not be from the most recent NeRF state. The use of iterative DU is aimed at maximizing the diversity of training ray viewpoints in each iteration, a choice that we find greatly improves both training stability and efficiency.
In the following section, we provide a comparison to a na\"ive adaptation of the SDS loss to our application.

\begin{figure}
    \centering
    \includegraphics[width=\linewidth]{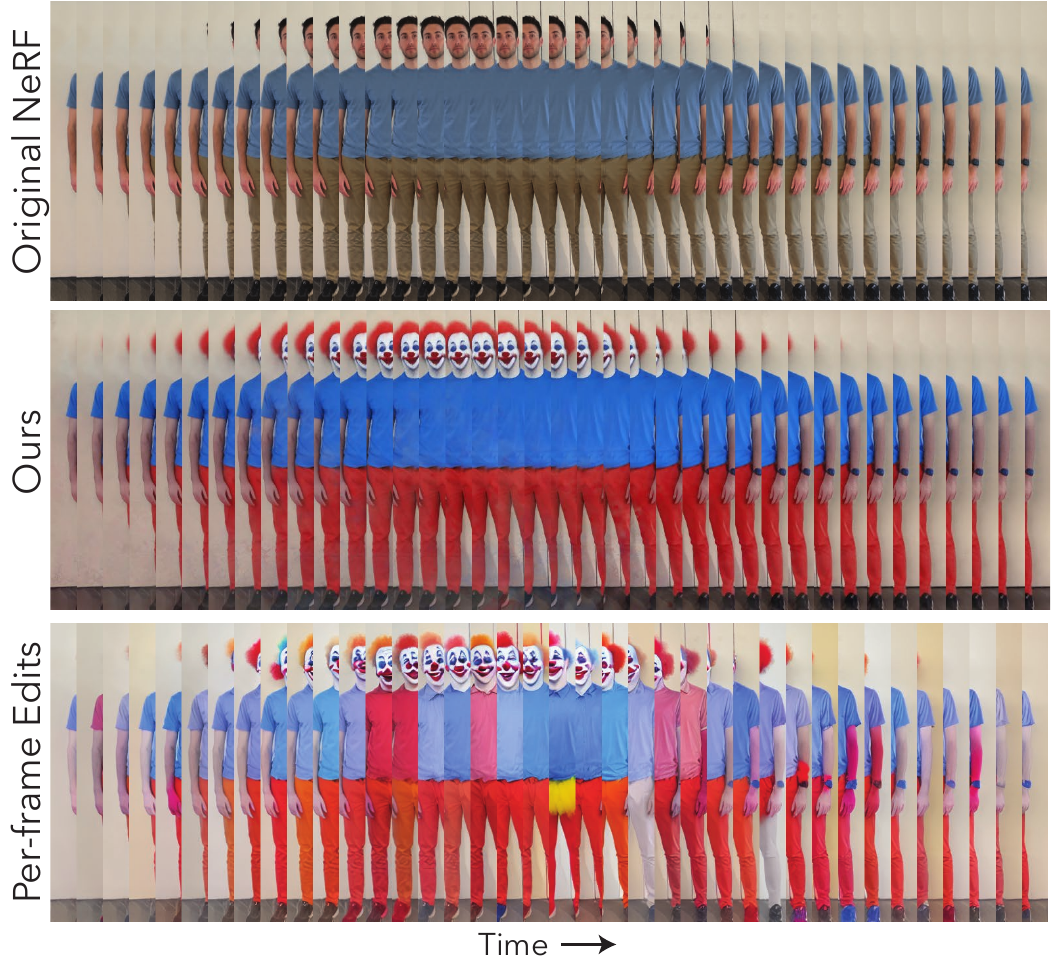}
    \caption{\textbf{Consistency:} Vertical slices of a rendered novel camera path show the consistency across varying viewpoints. The original NeRF rendering (top) is quite consistent, similar to our edited result (middle), using the prompt \textit{``turn him into a clown"}. Conversely, running InstructPix2Pix~\cite{brooks2022instructpix2pix} on each rendered frame independently results in notable inconsistency, such as varying hair and shirt colors.}
    \label{fig:consistency}
\end{figure}
\begin{figure*}
    \centering
    \includegraphics[width=\linewidth]{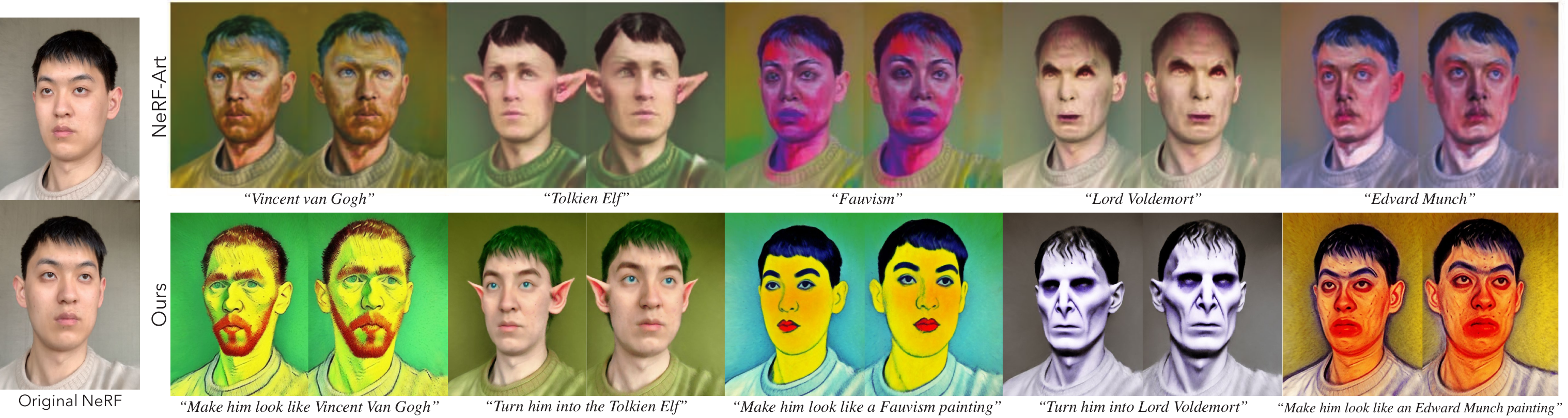}
    \caption{\textbf{Comparison with NeRF-Art}: We compare with CLIP-based method NeRF-Art on sequences and edits provided in their paper.}
    \label{fig:nerfart-comp}
\end{figure*}
\begin{figure*}
    \centering
    \includegraphics[width=\linewidth]{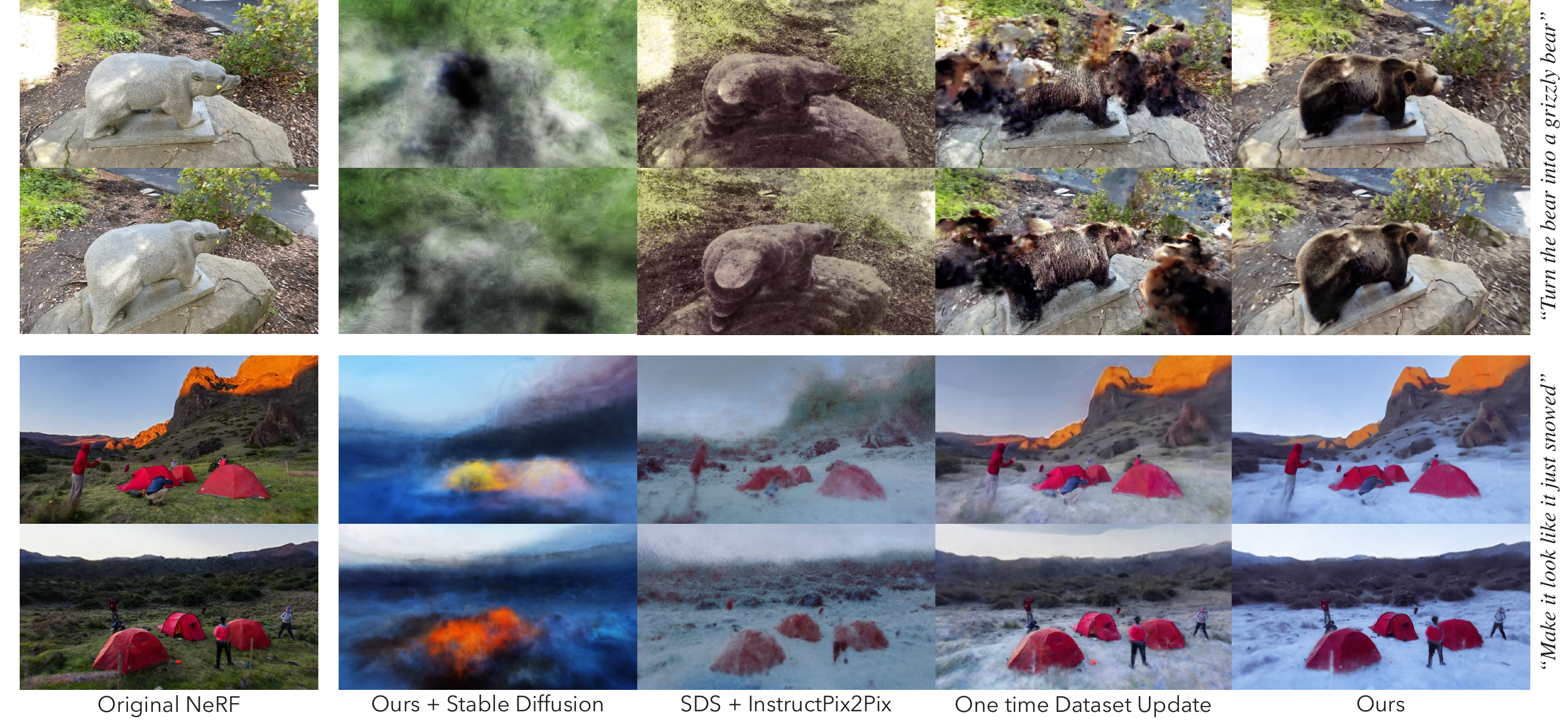}
    \caption{\textbf{Baseline Comparisons:} We compare our model with a collection of variants described in Section~\ref{sec:ablations}.}
    \label{fig:baselines}
\end{figure*}

\subsection{Implementation details}
As the underlying NeRF implementation, we use the `nerfacto' model from NeRFStudio~\cite{nerfstudio}. The strength and consistency of the updates performed by the diffusion model are determined by several parameters. Among these are the values for $[t_\text{min}, t_\text{max}] = [0.02,0.98]$, which define the amount of noise (and therefore the amount signal retained from the original images). Regardless of $t$, we always sample our denoised image with 20 denoising steps. The diffusion model has additional parameters, such as the classifier-free guidance weights corresponding to the text and image conditioning signals. For these, we can use the default values of $s_I=1.5$ and $s_T=7.5$, or offer the user the ability to hand-tune the guidance weight on an image to achieve the optimal edit strength before performing our NeRF optimization process. The results shown in the paper use manually selected guidance values, but adjusting these can result in varying degrees of scene edits, as shown in Figure~\ref{fig:cfg_Scale}. All other 3D hyperparameters are fixed for all experiments. During optimization, for the sake of efficiency, we update one image at a time, i.e., $d=1$ and $n=10$. For NeRF training, we use L1 and LPIPS~\cite{zhang2018unreasonable} losses. We train our method for a maximum of 30k iterations, which takes roughly an hour on a single NVIDIA Titan RTX (15GB of memory). More details are provided in the appendix.

\section{Results}
We conduct experiments on real scenes optimized using Nerfstudio~\cite{nerfstudio}. We edit a variety of scenes that vary in complexity: 360 scenes of environments and objects, faces, and full-body portraits. 
The scenes were captured using both a smartphone and a mirrorless camera. The camera poses were extracted using either COLMAP~\cite{schonberger2016structure} or through the PolyCam~\cite{polycam} app. The size of each dataset ranges from 50-300 images. First, we evaluate our approach through a variety of qualitative evaluations. To validate our design choices, we compare against a set of ablative baselines both qualitatively and quantitatively. Additionally, we provide visual comparisons to concurrent work NeRF-Art~\cite{wang2022nerf}.

\subsection{Qualitative Evaluation}
\paragraph{Editing 3D Scenes}
Our qualitative results are shown in Figure~\ref{fig:teaser} and Figure~\ref{fig:main_results}. For each edit, we show multiple views to illustrate the 3D consistency. On the portrait capture in Figure~\ref{fig:teaser}, we are able to achieve a broad range of edits varying from global (\textit{``Turn him into a Modigliani painting''}) to locally specific edits (\textit{``Turn his face into a skull''}). Although adding a completely new object is as challenging as the task of DreamFusion, our approach is able to add contextual elements such as \textit{``Give him a cowboy hat''} and \textit{``mustache''}.
Moreover, our method is able to dress the person to some degree, such as those illustrated on the full-body portrait in Figure~\ref{fig:main_results}, third row. It can achieve material changes such as \textit{``As a bronze bust''} and \textit{``Make him a marble statue''}. In the \textit{``bronze''} cases a subtle amount of view-dependent changes are also captured.
Our approach is also able to turn portraits into notable figures such as Einstein and fictional characters like \textit{``Batman''}. These edits also extend to subjects other than people, like changing a bear statue into a real polar bear, panda, and grizzly bear (Figure~\ref{fig:main_results}, last row). Most notably, these edits also apply to large-scale scenes (Figure~\ref{fig:main_results}, first row, Figure~\ref{fig:baselines}, bottom), and support instructions that modify the time of the day, seasons, and other conditions such as snow and desert.

\paragraph{Ablation Study}
\label{sec:ablations}
We validate our design choices by comparing our approach to the following variants. The qualitative differences are shown in Figure~\ref{fig:baselines}: \\

\noindent \textit{Per-frame Edit.} Our most na\"ive baseline is to apply InstructPix2Pix~\cite{brooks2022instructpix2pix} independently on every frame of a novel path rendered by the original NeRF. 
We use the rendered images as $c_I$, and the same text instruction as $c_T$. For $z_t$, we use pure noise. Despite the fact that the conditioning images are 3D consistent, the resulting edited images have significant variance that is inconsistent across different views. We illustrate this inconsistency in Figure~\ref{fig:consistency}, where we pan a camera across the scene and concatenate a slice from each frame to create an image. See supplemental video for examples.

\noindent \textit{One time Dataset Update.} In this baseline, we perform a single Dataset Update step, in which all training images are edited once, and the NeRF is trained until convergence on those edited images. 
The quality of this baseline depends largely on the 3D consistency of the per-frame editing results. 
While this approach can sometimes yield decent results, in a majority of cases, the initial edited 2D images are largely inconsistent, leading to blurry and artifact-filled 3D scenes, as shown in Figure~\ref{fig:baselines}. This problem is even more prominent when contextual objects are added to the portraits.

\begin{table}
    \begin{center}
        \centering
        \scriptsize{
        \begin{tabular}{l|c|c}
            \toprule 
             & \scriptsize{CLIP Text-Image} & \scriptsize{CLIP Direction}\\ & \scriptsize{Direction Similarity} $\uparrow$ & \scriptsize{Consistency} $\uparrow$ \\ \midrule
             Per-frame IP2P~\cite{brooks2022instructpix2pix} & \textbf{0.1603} & 0.8185 \\
             One-time DU & 0.1157 & 0.8823 \\
             SDS w/ IP2P~\cite{brooks2022instructpix2pix} & 0.0266 & \textit{0.9160} \\
             Ours & \textit{0.1600} & \textbf{0.9191}\\
             \bottomrule
        \end{tabular}
        }
    \end{center}
    \caption{\textbf{Quantiative Evaluation.}
Although edits are subjective, we provide quantitative metrics that evaluate the alignment of the edits to the text and consistency between subsequent frames in the CLIP space. Our approach results in similar CLIP similarity as per-frame edit, while achieving best consistency in CLIP space.
    }
    \label{tab:metrics}
\end{table}

\noindent \textit{DreamFusion (text-conditioned diffusion).} The next approach is to naively apply DreamFusion optimization to an existing NeRF scene. Specifically, starting from a NeRF initialized by the density and appearance obtained from real images, we apply SDS loss~\cite{poole2022dreamfusion} using StableDiffusion~\cite{rombach2022high}, a text-only diffusion model. We observed that this method quickly diverges and thus we do not include qualitative results in the paper. The reason for this divergence is that, in this setting, every image needs a textual description of the scene, and it becomes difficult to find an exact textual description that matches a scene across all views, especially for those with 360-degree coverage. This experiment highlights the importance of image conditioning.

\noindent \textit{SDS + InstructPix2Pix.}
If instead, we use an image-conditioned generative model, InstructPix2Pix, with the SDS loss from the previous variant, we are able to circumvent the requirements for an accurate text description of the whole scene. Unlike the text-conditioned variant, this approach does not diverge, but results in a 3D scene with more artifacts, as seen in Figure~\ref{fig:baselines}, third column.
We largely attribute this to the fact that the standard SDS samples rays from a small collection of full images, which makes optimization more unreliable than sampling rays randomly across all viewpoints.

\noindent \textit{Ours + StableDiffusion.} Finally, we compare our approach (with Iterative DU), but using StableDiffusion instead of InstructPix2Pix. This approach suffers from similar issues as seen in the DreamFusion baseline, because of the lack of image conditioning. Although it doesn't diverge, the resulting scene is blurry, and the 3D density is not coherent. Qualitative results can be seen in Figure~\ref{fig:baselines}, first column.

\begin{figure}
    \centering
    \includegraphics[width=.9\linewidth]{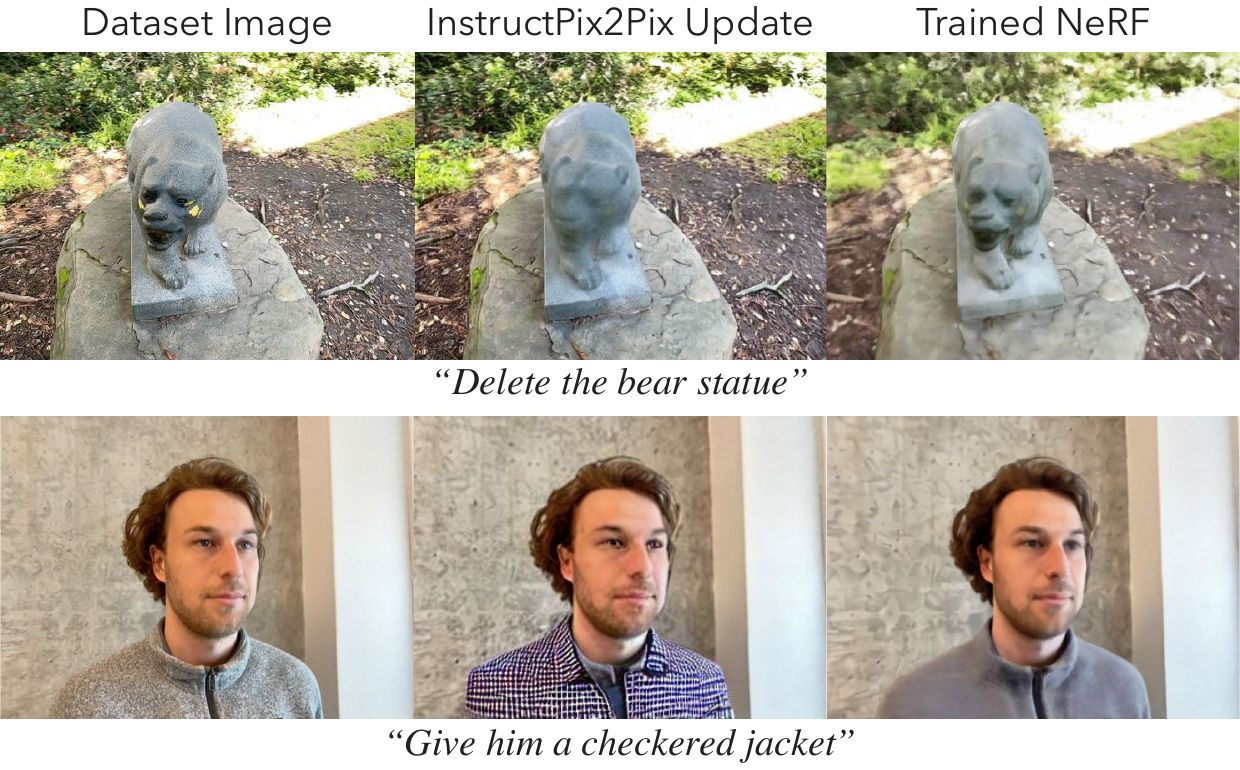}
    \caption{\textbf{Limitations:} InstructPix2Pix cannot always perform the desired edit (top), and thus our method does not perform an edit. Sometimes InstructPix2Pix produces correct, but inconsistent edits in 2D that our method fails to consolidate in 3D (bottom).}
    \label{fig:limitations}
\end{figure}

\paragraph{Comparisons with NeRF-Art.} We provide a qualitative comparison against concurrent work NeRF-Art~\cite{wang2022nerf}. Although their training code is unavailable, we use their provided custom-captured scenes and perform similar edits using our method. A comparison of their provided scene is shown in Figure~\ref{fig:nerfart-comp}. Note that their text inputs are not instructions, leaving the model with ambiguity on what exactly to edit. For instance, in their example of \textit{``Van Gogh''}, it's unclear whether the model should create a painting in the style of Van Gogh or make the face look like Van Gogh's face. Since edits are subjective, we leave it to the readers to determine their preference for these edits and provide this as a reference to a competitive state-of-the-art.

\subsection{Quantitative Evaluation}
Editing is fundamentally a subjective task. Thus, we mostly rely on various types of qualitative evaluation. We recommend the reader to evaluate the performance through the supplemental videos. Nevertheless, inspired by the evaluation protocols in InstructPix2Pix, we report auxiliary quantitative metrics over 10 total edits across two scenes, measuring (1) the alignment of the performed 3D edit with the text instruction (as shown in InstructPix2Pix and StyleGAN-Nada~\cite{gal2021stylegan} and (2) the temporal consistency of the performed edit across views, shown in Table \ref{tab:metrics}. The latter is a novel metric, similar to the CLIP directional similarity, but measuring the directional similarity between pairs of original and edited images in adjacent frames of novel rendered camera paths. More details are provided in the appendix.

\subsection{Limitations}

\looseness=-1 Our method inherits many of the limitations of InstructPix2Pix, such as the inability to perform large spatial manipulations.
Furthermore, as in DreamFusion, our method uses a diffusion model on a single view at a time, and thus may suffer from similar artifacts, such as double faces on added objects.
We demonstrate examples of two types of failure cases in Figure~\ref{fig:limitations}: (1) InstructPix2Pix fails to perform the edit in 2D, and therefore our method fails in 3D, and (2) InstructPix2Pix succeeds at editing in 2D, but has large inconsistencies that our method fails to consolidate in 3D. 

Specifically, our method is effective at instruction-driven, contextual, large-scale edits, which includes 1) editing textures, 2) replacing objects, and 3) changing global properties of a scene, among others. However, adding entirely new objects to the scene (such as adding a cup on a table) is challenging for various reasons. InstructPix2Pix struggles to add content into empty regions, and even when it is successful, it often places objects in different locations in different images, making 3D reconstruction a challenge. Similarly, InstructPix2Pix struggles to remove objects without replacing them with similarly salient (but also view-inconsistent) content, resulting in similar artifacts. We contend that as diffusion models improve at adding content, removing content, and overall image manipulation, our Iterative DU framework for NeRF editing will similarly improve. Limitations are further discussed in the appendix.

\section{Conclusion}

In this paper, we have introduced Instruct-NeRF2NeRF, a promising step towards the democratization of 3D scene editing for everyday users. Our method enables intuitive and accessible NeRF scene editing using natural text instructions. We operate on pre-captured NeRF scenes, ensuring that any resulting edits maintain 3D-consistency. We showed our method's results on a variety of captured NeRF scenes and demonstrated its ability to accomplish a wide range of edits on people, objects, and large-scale scenes.

\section{Acknowledgements}

We thank our colleagues for their insightful feedback helpful discussions, in particular Ethan Weber, Frederik Warburg, Ben Poole, Richard Szeliski, Jon Barron, Alexander Kristoffersen, Rohan Mathur, Alejandro Escontrela, and the Nerfstudio team.

{\small
\bibliographystyle{ieee_fullname}
\bibliography{main}

\begin{thebibliography}{10}\itemsep=-1pt

\bibitem{boss2021neuralpil}
Mark Boss, Varun Jampani, Raphael Braun, Ce Liu, Jonathan~T. Barron, and
  Hendrik~P.A. Lensch.
\newblock Neural-pil: Neural pre-integrated lighting for reflectance
  decomposition.
\newblock In {\em Advances in Neural Information Processing Systems (NeurIPS)},
  2021.

\bibitem{brooks2022instructpix2pix}
Tim Brooks, Aleksander Holynski, and Alexei~A. Efros.
\newblock Instructpix2pix: Learning to follow image editing instructions.
\newblock In {\em CVPR}, 2023.

\bibitem{brown2020language}
Tom Brown, Benjamin Mann, Nick Ryder, Melanie Subbiah, Jared~D Kaplan, Prafulla
  Dhariwal, Arvind Neelakantan, Pranav Shyam, Girish Sastry, Amanda Askell,
  et~al.
\newblock Language models are few-shot learners.
\newblock {\em Advances in neural information processing systems},
  33:1877--1901, 2020.

\bibitem{caron2021emerging}
Mathilde Caron, Hugo Touvron, Ishan Misra, Herv{\'e} J{\'e}gou, Julien Mairal,
  Piotr Bojanowski, and Armand Joulin.
\newblock Emerging properties in self-supervised vision transformers.
\newblock In {\em Proceedings of the IEEE/CVF international conference on
  computer vision}, pages 9650--9660, 2021.

\bibitem{chiang2021stylizing}
Pei-Ze Chiang, Meng-Shiun Tsai, Hung-Yu Tseng, Wei sheng Lai, and Wei-Chen
  Chiu.
\newblock Stylizing 3d scene via implicit representation and hypernetwork,
  2021.

\bibitem{gal2021stylegan}
Rinon Gal, Or Patashnik, Haggai Maron, Gal Chechik, and Daniel Cohen-Or.
\newblock Stylegan-nada: Clip-guided domain adaptation of image generators.
\newblock {\em arXiv preprint arXiv:2108.00946}, 2021.

\bibitem{gatys2016image}
Leon~A Gatys, Alexander~S Ecker, and Matthias Bethge.
\newblock Image style transfer using convolutional neural networks.
\newblock In {\em Proceedings of the IEEE conference on computer vision and
  pattern recognition}, pages 2414--2423, 2016.

\bibitem{hertzmann1998painterly}
Aaron Hertzmann.
\newblock Painterly rendering with curved brush strokes of multiple sizes.
\newblock In {\em Proceedings of the 25th annual conference on Computer
  graphics and interactive techniques}, pages 453--460, 1998.

\bibitem{ho2020denoising}
Jonathan Ho, Ajay Jain, and Pieter Abbeel.
\newblock Denoising diffusion probabilistic models.
\newblock {\em Advances in Neural Information Processing Systems},
  33:6840--6851, 2020.

\bibitem{hong2022avatarclip}
Fangzhou Hong, Mingyuan Zhang, Liang Pan, Zhongang Cai, Lei Yang, and Ziwei
  Liu.
\newblock Avatarclip: Zero-shot text-driven generation and animation of 3d
  avatars.
\newblock {\em ACM Transactions on Graphics (TOG)}, 41(4):1--19, 2022.

\bibitem{huang23vlmaps}
Chenguang Huang, Oier Mees, Andy Zeng, and Wolfram Burgard.
\newblock Visual language maps for robot navigation.
\newblock In {\em Proceedings of the IEEE International Conference on Robotics
  and Automation (ICRA)}, London, UK, 2023.

\bibitem{huang_2021_3d_scene_stylization}
Hsin-Ping Huang, Hung-Yu Tseng, Saurabh Saini, Maneesh Singh, and Ming-Hsuan
  Yang.
\newblock Learning to stylize novel views.
\newblock {\em arXiv preprint arXiv:2105.13509}, 2021.

\bibitem{huang2022stylizednerf}
Yi-Hua Huang, Yue He, Yu-Jie Yuan, Yu-Kun Lai, and Lin Gao.
\newblock Stylizednerf: consistent 3d scene stylization as stylized nerf via
  2d-3d mutual learning.
\newblock In {\em Proceedings of the IEEE/CVF Conference on Computer Vision and
  Pattern Recognition}, pages 18342--18352, 2022.

\bibitem{jain2021dreamfields}
Ajay Jain, Ben Mildenhall, Jonathan~T. Barron, Pieter Abbeel, and Ben Poole.
\newblock Zero-shot text-guided object generation with dream fields.
\newblock {\em CVPR}, 2022.

\bibitem{kobayashi2022distilledfeaturefields}
Sosuke Kobayashi, Eiichi Matsumoto, and Vincent Sitzmann.
\newblock Decomposing nerf for editing via feature field distillation.
\newblock In {\em Advances in Neural Information Processing Systems},
  volume~35, 2022.

\bibitem{lee2022understanding}
Han-Hung Lee and Angel~X Chang.
\newblock Understanding pure clip guidance for voxel grid nerf models.
\newblock {\em arXiv preprint arXiv:2209.15172}, 2022.

\bibitem{li2022language}
Boyi Li, Kilian~Q Weinberger, Serge Belongie, Vladlen Koltun, and Ren{\'e}
  Ranftl.
\newblock Language-driven semantic segmentation.
\newblock {\em arXiv preprint arXiv:2201.03546}, 2022.

\bibitem{li2022climatenerf}
Yuan Li, Zhi-Hao Lin, David Forsyth, Jia-Bin Huang, and Shenlong Wang.
\newblock Climatenerf: Physically-based neural rendering for extreme climate
  synthesis.
\newblock {\em arXiv e-prints}, pages arXiv--2211, 2022.

\bibitem{lin2022magic3d}
Chen-Hsuan Lin, Jun Gao, Luming Tang, Towaki Takikawa, Xiaohui Zeng, Xun Huang,
  Karsten Kreis, Sanja Fidler, Ming-Yu Liu, and Tsung-Yi Lin.
\newblock Magic3d: High-resolution text-to-3d content creation.
\newblock {\em arXiv preprint arXiv:2211.10440}, 2022.

\bibitem{liu2021editing}
Steven Liu, Xiuming Zhang, Zhoutong Zhang, Richard Zhang, Jun-Yan Zhu, and
  Bryan Russell.
\newblock Editing conditional radiance fields.
\newblock In {\em Proceedings of the International Conference on Computer
  Vision (ICCV)}, 2021.

\bibitem{melas2023realfusion}
Luke Melas-Kyriazi, Christian Rupprecht, Iro Laina, and Andrea Vedaldi.
\newblock Realfusion: 360° reconstruction of any object from a single image.
\newblock {\em arXiv e-prints}, pages arXiv--2302, 2023.

\bibitem{meng2021sdedit}
Chenlin Meng, Yutong He, Yang Song, Jiaming Song, Jiajun Wu, Jun-Yan Zhu, and
  Stefano Ermon.
\newblock Sdedit: Guided image synthesis and editing with stochastic
  differential equations.
\newblock In {\em International Conference on Learning Representations}, 2021.

\bibitem{metzer2022latent}
Gal Metzer, Elad Richardson, Or Patashnik, Raja Giryes, and Daniel Cohen-Or.
\newblock Latent-nerf for shape-guided generation of 3d shapes and textures.
\newblock {\em arXiv preprint arXiv:2211.07600}, 2022.

\bibitem{text2mesh}
Oscar Michel, Roi Bar-On, Richard Liu, Sagie Benaim, and Rana Hanocka.
\newblock Text2mesh: Text-driven neural stylization for meshes.
\newblock {\em arXiv preprint arXiv:2112.03221}, 2021.

\bibitem{mildenhall2022rawnerf}
Ben Mildenhall, Peter Hedman, Ricardo Martin-Brualla, Pratul~P. Srinivasan, and
  Jonathan~T. Barron.
\newblock {NeRF} in the dark: High dynamic range view synthesis from noisy raw
  images.
\newblock {\em CVPR}, 2022.

\bibitem{mildenhall2020nerf}
Ben Mildenhall, Pratul~P. Srinivasan, Matthew Tancik, Jonathan~T. Barron, Ravi
  Ramamoorthi, and Ren Ng.
\newblock Nerf: Representing scenes as neural radiance fields for view
  synthesis.
\newblock In {\em ECCV}, 2020.

\bibitem{munkberg2022extracting}
Jacob Munkberg, Jon Hasselgren, Tianchang Shen, Jun Gao, Wenzheng Chen, Alex
  Evans, Thomas M{\"u}ller, and Sanja Fidler.
\newblock Extracting triangular 3d models, materials, and lighting from images.
\newblock In {\em Proceedings of the IEEE/CVF Conference on Computer Vision and
  Pattern Recognition}, pages 8280--8290, 2022.

\bibitem{nguyen2022snerf}
Thu Nguyen-Phuoc, Feng Liu, and Lei Xiao.
\newblock Snerf: stylized neural implicit representations for 3d scenes.
\newblock {\em ACM Transactions on Graphics (TOG)}, 41(4):1--11, 2022.

\bibitem{chatgpt}
OpenAI.
\newblock {ChatGPT}.

\bibitem{ost2021neural}
Julian Ost, Fahim Mannan, Nils Thuerey, Julian Knodt, and Felix Heide.
\newblock Neural scene graphs for dynamic scenes.
\newblock In {\em Proceedings of the IEEE/CVF Conference on Computer Vision and
  Pattern Recognition}, pages 2856--2865, 2021.

\bibitem{ouyang2022training}
Long Ouyang, Jeff Wu, Xu Jiang, Diogo Almeida, Carroll~L Wainwright, Pamela
  Mishkin, Chong Zhang, Sandhini Agarwal, Katarina Slama, Alex Ray, et~al.
\newblock Training language models to follow instructions with human feedback.
\newblock {\em arXiv preprint arXiv:2203.02155}, 2022.

\bibitem{polycam}
Polycam.
\newblock Polycam - lidar \& 3d scanner for iphone \& android.

\bibitem{poole2022dreamfusion}
Ben Poole, Ajay Jain, Jonathan~T. Barron, and Ben Mildenhall.
\newblock Dreamfusion: Text-to-3d using 2d diffusion.
\newblock {\em arXiv}, 2022.

\bibitem{radford2021learning}
Alec Radford, Jong~Wook Kim, Chris Hallacy, Aditya Ramesh, Gabriel Goh,
  Sandhini Agarwal, Girish Sastry, Amanda Askell, Pamela Mishkin, Jack Clark,
  et~al.
\newblock Learning transferable visual models from natural language
  supervision.
\newblock In {\em International conference on machine learning}, pages
  8748--8763. PMLR, 2021.

\bibitem{ramesh2022hierarchical}
Aditya Ramesh, Prafulla Dhariwal, Alex Nichol, Casey Chu, and Mark Chen.
\newblock Hierarchical text-conditional image generation with clip latents.
\newblock {\em arXiv preprint arXiv:2204.06125}, 2022.

\bibitem{rombach2022high}
Robin Rombach, Andreas Blattmann, Dominik Lorenz, Patrick Esser, and Bj{\"o}rn
  Ommer.
\newblock High-resolution image synthesis with latent diffusion models.
\newblock In {\em Proceedings of the IEEE/CVF Conference on Computer Vision and
  Pattern Recognition}, pages 10684--10695, 2022.

\bibitem{saharia2022photorealistic}
Chitwan Saharia, William Chan, Saurabh Saxena, Lala Li, Jay Whang, Emily
  Denton, Seyed Kamyar~Seyed Ghasemipour, Burcu~Karagol Ayan, S~Sara Mahdavi,
  Rapha~Gontijo Lopes, et~al.
\newblock Photorealistic text-to-image diffusion models with deep language
  understanding.
\newblock {\em arXiv preprint arXiv:2205.11487}, 2022.

\bibitem{schonberger2016structure}
Johannes~L Schonberger and Jan-Michael Frahm.
\newblock Structure-from-motion revisited.
\newblock In {\em Proceedings of the IEEE conference on computer vision and
  pattern recognition}, pages 4104--4113, 2016.

\bibitem{sohl2015deep}
Jascha Sohl-Dickstein, Eric Weiss, Niru Maheswaranathan, and Surya Ganguli.
\newblock Deep unsupervised learning using nonequilibrium thermodynamics.
\newblock In {\em International Conference on Machine Learning}, pages
  2256--2265. PMLR, 2015.

\bibitem{nerv2021}
Pratul~P. Srinivasan, Boyang Deng, Xiuming Zhang, Matthew Tancik, Ben
  Mildenhall, and Jonathan~T. Barron.
\newblock Nerv: Neural reflectance and visibility fields for relighting and
  view synthesis.
\newblock In {\em CVPR}, 2021.

\bibitem{nerfstudio}
Matthew Tancik, Ethan Weber, Evonne Ng, Ruilong Li, Brent Yi, Justin Kerr,
  Terrance Wang, Alexander Kristoffersen, Jake Austin, Kamyar Salahi, Abhik
  Ahuja, David McAllister, and Angjoo Kanazawa.
\newblock Nerfstudio: A modular framework for neural radiance field
  development.
\newblock {\em arXiv preprint arXiv:2302.04264}, 2023.

\bibitem{tewari2022advances}
Ayush Tewari, Justus Thies, Ben Mildenhall, Pratul Srinivasan, Edgar Tretschk,
  Wang Yifan, Christoph Lassner, Vincent Sitzmann, Ricardo Martin-Brualla,
  Stephen Lombardi, et~al.
\newblock Advances in neural rendering.
\newblock In {\em Computer Graphics Forum}. Wiley Online Library, 2022.

\bibitem{tschernezki22neural}
Vadim Tschernezki, Iro Laina, Diane Larlus, and Andrea Vedaldi.
\newblock {Neural Feature Fusion Fields}: {3D} distillation of self-supervised
  {2D} image representations.
\newblock In {\em Proceedings of the International Conference on {3D} Vision
  (3DV)}, 2022.

\bibitem{verbin2022refnerf}
Dor Verbin, Peter Hedman, Ben Mildenhall, Todd Zickler, Jonathan~T. Barron, and
  Pratul~P. Srinivasan.
\newblock {Ref-NeRF}: Structured view-dependent appearance for neural radiance
  fields.
\newblock {\em CVPR}, 2022.

\bibitem{von-platen-etal-2022-diffusers}
Patrick von Platen, Suraj Patil, Anton Lozhkov, Pedro Cuenca, Nathan Lambert,
  Kashif Rasul, Mishig Davaadorj, and Thomas Wolf.
\newblock Diffusers: State-of-the-art diffusion models.
\newblock \url{https://github.com/huggingface/diffusers}, 2022.

\bibitem{wang2021clip}
Can Wang, Menglei Chai, Mingming He, Dongdong Chen, and Jing Liao.
\newblock Clip-nerf: Text-and-image driven manipulation of neural radiance
  fields.
\newblock {\em arXiv preprint arXiv:2112.05139}, 2021.

\bibitem{wang2022nerf}
Can Wang, Ruixiang Jiang, Menglei Chai, Mingming He, Dongdong Chen, and Jing
  Liao.
\newblock Nerf-art: Text-driven neural radiance fields stylization.
\newblock {\em arXiv preprint arXiv:2212.08070}, 2022.

\bibitem{wang2022score}
Haochen Wang, Xiaodan Du, Jiahao Li, Raymond~A Yeh, and Greg Shakhnarovich.
\newblock Score jacobian chaining: Lifting pretrained 2d diffusion models for
  3d generation.
\newblock {\em arXiv preprint arXiv:2212.00774}, 2022.

\bibitem{wu2022palettenerf}
Qiling Wu, Jianchao Tan, and Kun Xu.
\newblock Palettenerf: Palette-based color editing for nerfs.
\newblock {\em arXiv preprint arXiv:2212.12871}, 2022.

\bibitem{orf}
Hong-Xing Yu, Leonidas~J Guibas, and Jiajun Wu.
\newblock Unsupervised discovery of object radiance fields.
\newblock {\em arXiv preprint arXiv:2107.07905}, 2021.

\bibitem{yuan2022nerf}
Yu-Jie Yuan, Yang-Tian Sun, Yu-Kun Lai, Yuewen Ma, Rongfei Jia, and Lin Gao.
\newblock Nerf-editing: geometry editing of neural radiance fields.
\newblock In {\em Proceedings of the IEEE/CVF Conference on Computer Vision and
  Pattern Recognition}, pages 18353--18364, 2022.

\bibitem{zhang2021editable}
Jiakai Zhang, Xinhang Liu, Xinyi Ye, Fuqiang Zhao, Yanshun Zhang, Minye Wu,
  Yingliang Zhang, Lan Xu, and Jingyi Yu.
\newblock Editable free-viewpoint video using a layered neural representation.
\newblock {\em ACM Transactions on Graphics (TOG)}, 40(4):1--18, 2021.

\bibitem{zhang2022arf}
Kai Zhang, Nick Kolkin, Sai Bi, Fujun Luan, Zexiang Xu, Eli Shechtman, and Noah
  Snavely.
\newblock Arf: Artistic radiance fields, 2022.

\bibitem{zhang2018unreasonable}
Richard Zhang, Phillip Isola, Alexei~A Efros, Eli Shechtman, and Oliver Wang.
\newblock The unreasonable effectiveness of deep features as a perceptual
  metric.
\newblock In {\em Proceedings of the IEEE conference on computer vision and
  pattern recognition}, pages 586--595, 2018.

\bibitem{zhou2022sparsefusion}
Zhizhuo Zhou and Shubham Tulsiani.
\newblock Sparsefusion: Distilling view-conditioned diffusion for 3d
  reconstruction, 2022.

\end{thebibliography}
}

\clearpage
\appendix

\section{Additional implementation details}
The primary input to our method is a NeRF reconstruction of a real scene. This reconstruction is obtained using the `nerfacto' model from NeRFStudio~\cite{nerfstudio}, trained for $30,000$ iterations per scene. Before running Instruct-NeRF2NeRF, we re-initialize the optimizers in NeRFStudio. In order to specify edits, we use InstructPix2Pix~\cite{brooks2022instructpix2pix}, where users specify the classifier-free guidance weights in order to explore the desired amount of change for a given edit. In our method, we inherit this parameter. Below we list these chosen values for the sequences shown in the paper:

\begin{enumerate}
\item \emph{Fig. 4 Tree}: $s_I$=1.5, $s_T\in[7.5,12.5]$
\item \emph{Fig. 8 Tents}: $s_I$=1.5, $s_T$=10.0 
\item \emph{Fig. 4 Person}: $s_I\in[1.5,1.75]$, $s_T\in[6.5,7.5]$
\item \emph{Fig. 1}: $s_I$=1.5, $s_T\in[6.5,7.5]$
\item \emph{Fig. 4 Bear}: $s_I=1.5$, $s_T=6.5$
\item \emph{Fig. 7}: $s_I\in[1.3,1.5]$, $s_T\in[6.5,8.5]$
\end{enumerate}

Our process of optimizing for an edited NeRF uses a diffusion model as guidance, which can produce a collection of temporally varying images (i.e., varying over the course of optimization). As a result, the optimization process does not have a single convergence point, as standard NeRF optimization does, where all images are sufficiently well explained by the reconstructed model. Therefore, the edited NeRF also varies in type and strength of edit over the course of optimization, and one must select an iteration at which to terminate optimization and visualize the edited scene. In practice, the optimal choice for training length is a subjective decision --- a user may prefer more subtle or more extreme edits that are best found at different stages of training. The results in the paper are shown after a varying number of iterations, $3000-4000$ for smaller scenes (e.g., Fig.~1 and Fig.~4~bear, which both have under 100 images each), and $7000-8000$ for larger scenes (the remaining scenes, with over 200-300 images each).

For InstructPix2Pix, we use the public implementation included in the Diffusers library~\cite{von-platen-etal-2022-diffusers}. 

\section{Limitations}

As mentioned in the main paper, our method inherits many of the limitations of InstructPix2Pix. This includes (1) the inability to perform large spatial manipulations, (2) the occasional inability to perform binding between objects referred to in the instruction and the corresponding scene objects, and (3) adding or removing large objects. 
Furthermore, as in DreamFusion, our method uses a diffusion model on a single view at a time, and thus may suffer from similar artifacts, such as double faces on added objects. Finally, it's worth noting that the edit instructions provided to InstructPix2Pix are sometimes more relevant to certain views than others. For example, if the instruction is to \textit{``turn the man into a bear''}, not all views may prominently feature the man, and therefore, certain views may consistently produce less of an edit or no edit at all. Our framework can easily incorporate improvements made in the InstructPix2Pix and its follow up works.

\begin{figure}
    \centering
    \includegraphics[width=0.1\textwidth]{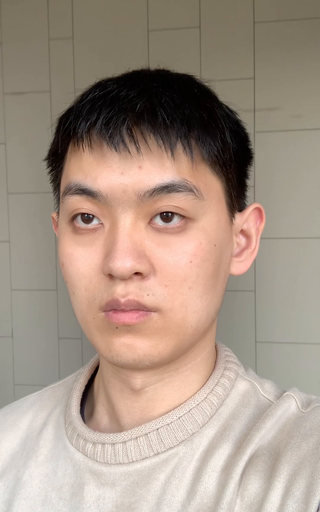}
    \includegraphics[width=0.1\textwidth]{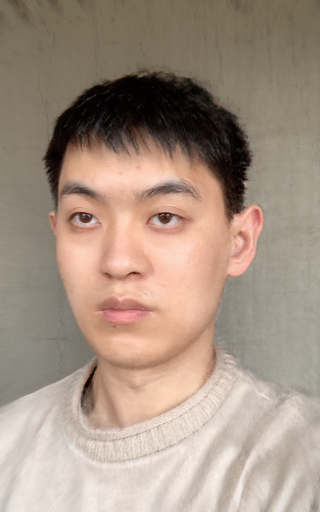}
    \includegraphics[width=0.1\textwidth]{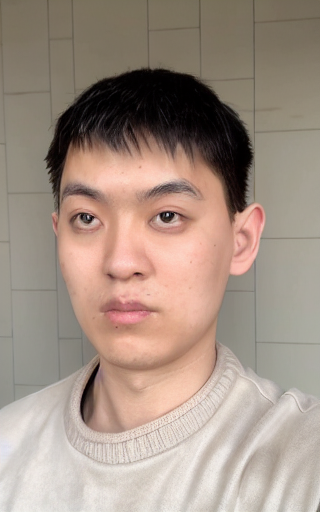}
    \includegraphics[width=0.1\textwidth]{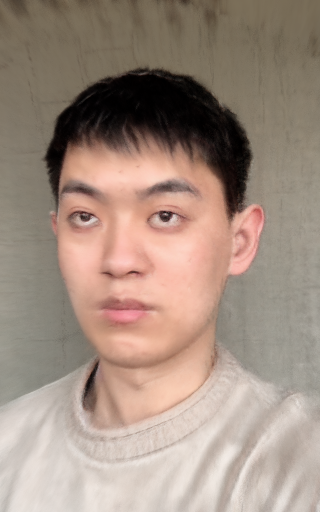}\\
    \includegraphics[width=0.1\textwidth,height=0.15\textwidth]{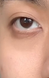}
    \includegraphics[width=0.1\textwidth,height=0.15\textwidth]{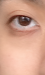}
    \includegraphics[width=0.1\textwidth,height=0.15\textwidth]{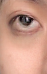}
    \includegraphics[width=0.1\textwidth,height=0.15\textwidth]{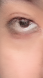}
    \caption{\textbf{Effects of autoencoder}: From left to right: (1) the input captured image, (2) the original reconstructed NeRF, rendered from the same viewpoint, (3), the original image 1, encoded and decoded by Stable Diffusion's autoencoder, (4) a NeRF reconstruction of the autoencoded images, rendered from the same viewpoint. Below, we show a zoomed-in patch of the eye. We note that while the input sequence can be reasonably reconstructed without much loss of detail, and the autoencoded images are similarly sharp as the input captured sequence, there are slight perturbations in local textures that end up not being 3D-consistent, resulting in blurry renders in the final recontruction. }
    \label{fig:blur}
\end{figure}

We additionally note that the edited NeRF scenes often contain slightly blurrier textures when compared to the originally reconstructed NeRF. Some initial experimentation suggests that this may be a result of the Stable Diffusion autoencoder, which often does not produce an entirely faithful copy of the original image, even in unedited regions. The autoencoder, while producing visually comparable results, often creates images with locally similar but non-identical textures that are not globally 3D-consistent. To validate this hypothesis, we show an experiment in Figure~\ref{fig:blur}, where we simply autoencode the input images using Stable Diffusion (i.e., the same autoencoder used by InstructPix2Pix) and continue training the NeRF. As a result, the scene becomes gradually blurrier. 

Furthermore, over much longer optimization runs (i.e., when the process of rendering, editing, and propagating the edited images back into the NeRF has been repeated many times), we note a decrease in visual quality. We also largely attribute this to the effects of the autoencoder. %

\section{Metrics}

In the quantitative evaluation, we report two metrics, a \textit{CLIP Directional Score}~\cite{gal2021stylegan}, and a metric we name \textit{CLIP Direction Consistency Score}. The CLIP Directional score measures how much the change in text captions agrees with the change in the images, and the CLIP Consistency score measures the cosine similarity of the CLIP embeddings of each pair of adjacent frames in a render novel camera path. 

More formally, for the directional score, we encode a pair of images (the original and edited NeRFs, rendered at a given viewpoint), as well as a pair of text prompts that describe the original and edited scenes, e.g., \textit{``a photograph of a man''} and \textit{``a photograph of a Tolkien Elf''}. Using these, we compute the directional score described in InstructPix2Pix~\cite{brooks2022instructpix2pix} and StyleGAN-NADA~\cite{gal2021stylegan}.

For the temporal consistency loss, we encode a pair of consecutive rendered frames from a novel trajectory, using both the original NeRF and the edited NeRF. In all, we are left with four CLIP embeddings, $C(o_i)$, $C(o_{i+1})$, $C(e_{i})$, $C(e_{i+1})$, corresponding to original NeRF renderings $o_{i}, o_{i+1}$ and edited NeRF renderings $e_{i}, e_{i+1}$ for consecutive novel views $i$ and $i+1$. We define the consistency loss as:
\begin{equation}
(C(e_i) - C(o_i))\cdot (C(e_{i+1}) - C(e_{i}))
\end{equation}
This effectively amounts to measing the change in the CLIP-space edit direction from frame to frame.

\end{document}